\documentclass[technote,a4paper,leqno]{IEEEtran}
\pdfoutput=1

\usepackage[utf8]{inputenc} 
\usepackage[russian,USenglish]{babel} 
\usepackage[T1]{fontenc}    
\usepackage{amsmath,amssymb}
\DeclareMathOperator{\ido}{ido}
\usepackage[absolute,overlay]{textpos}
\usepackage[dvipsnames]{xcolor}
\usepackage{tikz}
\usepackage{csquotes}
\usepackage[binary-units=true,detect-weight=true, detect-family=true]{siunitx}
\DeclareSIUnit\pixel{px}
\usepackage{caption}  
\usepackage{pbox}
\usepackage{pdfpages}
\usepackage{scrextend}

\usepackage{url}
\usepackage{breakurl}
\usepackage[raiselinks=true,
            bookmarks=true,
            bookmarksopenlevel=1,
            bookmarksopen=true,
            bookmarksnumbered=true,
            breaklinks,
            hyperindex=true,
            plainpages=false,
            pdfpagelabels=true,
            pdfborder={0 0 0.5}]{hyperref}

\usepackage{xspace}

\usepackage[nameinlink, noabbrev,capitalise]{cleveref}

\title{The WiLI benchmark dataset for written language identification}
\author{%
    \IEEEauthorblockN{Martin Thoma}\\
    \IEEEauthorblockA{E-Mail: info@martin-thoma.de} 
}

\hypersetup{
    pdfauthor   = {Martin Thoma},
    pdfkeywords = {dataset},
    pdfsubject  = {WiLI, dataset},
    pdftitle    = {The WiLI benchmark dataset for written natural language identification},
    pdflang     = {en-US},
    pdfdisplaydoctitle = true,
}
\usepackage[inline]{enumitem}
\usepackage{longtable}
\usepackage{booktabs}       
\usepackage{braket}         
\usepackage{algorithm,algpseudocode}

\usepackage{cjhebrew}
\usepackage{tipa}

\usepackage{parskip}
\usepackage{multirow}
\usepackage{microtype}

\newcommand{\dbTotalClasses}{235}
\newcommand{\dbTotalInstances}{\num{235000}}
\newcommand{\dbName}{WiLI}
\newcommand{\dbNameVersion}{WiLI-2018}
\newcommand{\dbSizeMB}{62.4}
\newcommand{\dbDownloadURL}{\url{https://doi.org/10.5281/zenodo.841984}}
\newcommand{\dbMDfivesum}{3dc5bd41587811ad6b0d04ae2f235f84}

\begin{document}
\maketitle
\begin{abstract}
This paper describes the \dbNameVersion{} benchmark dataset for monolingual
written natural language identification. \dbNameVersion{} is a publicly
available,\footnote{See appendix for detailed instructions how to obtain the
data.} free of charge dataset of short text extracts from Wikipedia. It
contains \num{1000}~paragraphs of \dbTotalClasses~languages, totaling in
\dbTotalInstances~paragraphs. \dbName{} is a classification dataset: Given an
unknown paragraph written in one dominant language, it has to be decided which
language it is.
\end{abstract}

\section{Introduction}
The identification of written natural language is a task which appears often in
web applications. Search engines want to provide users websites which are
relevant to them. Content which the users can't understand is automatically
less relevant. It is crucial for machine translation, sentiment analysis and
text summarization algorithms to know the source language. In the case of
document classification with documents of multiple languages one could train
one classifier per language, but to do so a reliable language identification is
necessary. OCR systems improve their recognition by applying knowledge of the
language of the scanned document.

While a couple of publications are available for written natural language
identification\cite{hughes2006reconsidering,tran2005markov,baldwin2010language,lui2014automatic},
none makes the used benchmark dataset publicly available and easy to access.
Publicly available datasets are important to make research reproducible. They
allow the community to analyze problems systematically and provide a
possibility for fair comparison of available solutions.

\dbName{} can be used to train models for language identification, for
benchmarking those models and for identifying unknown languages.

It is created with data from the free encyclopedia Wikipedia. Hence it is
expected that models which achieve good results on \dbName{} do not necessarily
achieve similar results on colloquial language document such as Twitter or
Facebook posts. Also, the dataset contains mostly natural language. This
includes constructed languages such as Esperanto, Ido, Interlingua and also
dead languages like Latin, but excludes artificial languages such as HTML and
XML, \LaTeX{}, JSON and Markdown.

Also, \dbNameVersion{} has exactly the same languages in the training set as in
the test set. So it is not necessary to predict unknown languages. In a real
setting, unknown languages can appear.

\section{Written Language Basics}
A language is a system of communication. This could be spoken language, written
language or sign language. A spoken language can have several
forms of written language. For example, the spoken Chinese language has at
least three writing systems: Traditional Chinese characters, Simplified Chinese
characters and Pinyin --- the official romanization system for Standard Chinese.

Languages evolve. New words like \textit{googling}, \textit{television}
and \textit{Internet} get added, but written languages are also refactored. For
example, the German orthography reform of 1996 aimed at making the written
language simpler. This means any system which recognizes language and any
benchmark needs to be adapted over time. Hence \dbName{} is versioned by year.

Languages do not necessarily have only one name. According to Wikipedia, the
Sranan language is also known as \textit{Sranan Tongo}, \textit{Sranantongo},
\textit{Surinaams}, \textit{Surinamese}, \textit{Surinamese Creole} and
\textit{Taki Taki}. This makes \mbox{ISO~369-3} valuable, but not all
languages are represented in \mbox{ISO~369-3}. As \mbox{ISO~369-3} uses
combinations of 3~Latin letters and has 547~reserved combinations, it can at
most represent \num{17029}~languages. As of January~2018, \mbox{ISO~369-3}
contains \num{7850}~languages~\cite{iso-369-3}.

Unicode is a standard for encoding text. A Unicode code point is an integer
which uniquely defines a single character. As of January 2018, Unicode~10.0
defines \num{136755}~characters. This includes the letters of 139~scripts such
as the Latin, Arabic, Greek, Cyrillic and Han script. Unicode~10.0 also
includes Emoji and other symbol characters such as currency symbols.

The most basic building block is a character. For example, \textit{ABCabc} are
six characters from the Latin script, \foreignlanguage{russian}{Текная} are six
characters from the Cyrillic script, and \begin{cjhebrew}br+s\end{cjhebrew} are
four Unicode code points from Hebrew script and two directional characters:
\texttt{\textbackslash{}u202b \textbackslash{}u05c1 \textbackslash{}u05e9
\textbackslash{}u05e8 \textbackslash{}u05d1 \textbackslash{}u202c}. The term
\textit{character} is used interchangeably with Unicode code points in this
paper. The length of a piece of text is defined by the number of Unicode code
points in it.

Unicode has combining characters and also the combination of some. For example,
the combination of \texttt{U+006E} (n) and \texttt{U+0303} ($\sim$) form ñ,
similar to the single Unicode code point \texttt{U+00F1} (ñ). To prevent such
differences, the Unicode normalization form C is applied~\cite{MarkDavis2017}:
Canonical Decomposition, followed by Canonical Composition.

Although this paper is about the identification of language solely by examining
the given text, there are many ways to get indicators for the language by
metadata. Such examples are geolocation and Website tags:
\begin{itemize}
    \small
    \item \small\texttt{<meta name="language" content="Spanish">}
    \item \small\texttt{<meta http-equiv="content-language" content="es">}
    \item \small\texttt{<html lang="es">}
    \item \small\texttt{<link rel="alternate" href="http://example.com/en-au" hreflang="en-au" />}
\end{itemize}

\section{Benchmark Datasets for LID}
The ALTW~2010~dataset~\cite{vrl2010multilingual} makes use of different
language Wikipedias. It artificially concatenates sections of articles in
different languages. This dataset was created to benchmark multilingual
language identification for 74~languages. This dataset is publicly available at
\url{http://people.eng.unimelb.edu.au/tbaldwin/etc/altw2010-langid.tgz}.

The \textit{Language Identification in Code-Switched (CS) data} shared
task~\cite{solorio2014overview} is a dataset which consists of four language
pairs (Modern Standard Arabic -- Dialectal Arabic, Mandarin -- English, Nepali
-- English, Spanish--English) for which data was collected mostly from Twitter.
The task is to label each token with one of five labels: \textit{lang1},
\textit{lang2}, \textit{other}, \textit{ambiguous}, \textit{mixed} or
\textit{named entity}. In this case, other refers to emoticons, punctuation
marks and numbers, ambiguous to words which belong to both language and
mixed to words which are a mixture of both languages. This dataset only
covers language pairs and only has four pairs.

The VarDial Workshop~2014 had a shared task for \textit{discriminating between
similar languages and language varieties} (DSL)~\cite{MarcosZampieri2014}.
There were six groups of languages:
\begin{table}[H]
    \begin{tabular}{llrl}
    \toprule
    ~ & Group                                     & Best accuracy & (Team) \\\midrule
    A & Bosnian vs Croatian vs Serbian            & \SI{93.60}{\percent} &(NRC-CNRC)    \\
    B & Indonesian vs Malaysian                   & \SI{99.55}{\percent} &(NRC-CNRC)    \\
    C & Czech vs Slovakian                        & \SI{100}{\percent}  &(NRC-CNRC)      \\
    D & Brazilian vs European Portuguese          & \SI{95.60}{\percent} &(NRC-CNRC)    \\
    E & Peninsular vs Argentine Spanish           & \SI{90.95}{\percent} &(NRC-CNRC)    \\
    F & American vs British English               & \SI{63.94}{\percent} &(UMich)       \\\bottomrule
    \end{tabular}
    \caption{Results of the VarDial Workshop 2014 for the similar language discrimination task.}
    \label{table:var-dial-2014-comparison}
\end{table}
DSL is important for spell-checking, for native language identification and for
translation. For example, \textit{pants} in British English means underwear
while it means trousers in American English. The VarDial Workshop~2014 dataset
is publicly available at \url{http://ttg.uni-saarland.de/resources/DSLCC/}.

TweetLID~\cite{zubiaga2016tweetlid} is a dataset of Tweets. It contains
\num{14992} Tweets for training and \num{19993} Tweets for testing. There are
10~classes for the dataset: Spanish, 
Portuguese, 
Catalan, 
English, 
Galician, 
Basque, 
Undeterm., 
Multilingual, 
Ambiguous, 
and Other. 
So the dataset only distinguishes 6~languages.

Another idea how to evaluate language identification is to take Machine
Translation dataset. For example, Europarl-v7 contains 21~languages and
60~million words per language~\cite{koehn2005europarl}.

\dbNameVersion{} covers more languages than any of those datasets. It is
balanced and easier to obtain (see Appendix).

\section{How \dbName{} was created}
\dbName{} consists of paragraphs from different language Wikipedias. The
\textit{Random page} feature was used via the Python \texttt{wikipedia}
package~\cite{Goldsmith2014} to obtain a page. The markup was stripped, the
text was split at headers and at
newlines. Leading and trailing whitespace characters was removed from the result and
consecutive whitespace characters were replaced by a single whitespace. The Unicode
normalization form~C was applied. If the resulting items length was at least
140~Unicode Code points and if it did not have the sequences \texttt{ISBN},
\verb+\displaystyle+, \texttt{vol. [digit of roman number]}
nor a valid DOI in it, then it was added to the list of paragraphs. This
process was repeated until 1000~paragraphs were collected.

The reason for checking for ISBN and DOI is that all Wikipedias contain many
references. The language of the reference is often English, even in non-English
Wikipedias.

\section{Classes}
The \dbNameVersion{}~dataset contains \dbTotalClasses~classes. Those classes
include 122~Indo-European languages, 22~Austronesian languages, 17~Turkic
languages, 14~Uralic languages, 11~Niger-Congo languages, 10~Sino-Tibetan languages,
9~Afro-Asiatic languages, 6~constructed languages and 24~languages of smaller
families.

For a couple of languages, there are variants available and languages which are
close to each other:
\begin{itemize}
    \item Arabic, Egyptian Arabic
    \item English, Old English, Scots
    \item Standard Chinese, Min Nan Chinese, Hakka Chinese, Literary Chinese, Wu Chinese
    \item German, Bavarian, Low German, Palatine German, Ripuarisch, Alemannic German, Pennsylvania German
    \item Belarusian, Belarusian (Taraschkewiza)
    \item Kurdish, Central Kurdish
    \item Indonesian, Minangkabau, Banyumasan, Banjar, Sundanese, Javanese
    \item Languages spoken in India:
    \begin{itemize}
        \item Maithili, Bhojpuri
        \item Bengali, Bishnupriya
        \item Konkani, Marathi
    \end{itemize}
    \item Russian, Komi-Permyak
    \item Persian, Gilaki, Mazanderani
\end{itemize}

See \cref{sec:languages} for a list of all \dbTotalClasses{}~languages within
\dbNameVersion.

\section{Data}
The \dbNameVersion{}~dataset contains \dbTotalInstances{}~paragraphs of at least
140~Unicode code points. Each paragraph belongs to one of
\dbTotalClasses{}~languages.

While the minimum paragraph length is guaranteed to be 140 and each languages
shortest paragraph is at most 144~Unicode code points long, the mean
paragraph length averaged over all languages is \num{371}~characters and
the longest paragraph is \num{195402}~Unicode code points long.

Each language can contain short excerpts from other languages and special characters.
For example, the English Wikipedia article about the Russian mathematician
Pafnuty Chebyshev starts like this:
\begin{displayquote}
Pafnuty Lvovich Chebyshev (Russian: \foreignlanguage{russian}{Пафнyтий Львoвич Чебышёв}; IPA: [p\textturna{}f\textprimstress{}nut\textsuperscript{j}\textsci{}j \textprimstress{}l\textsuperscript{j}vov\textsuperscript{j}\textsci{}\textipa{tc tc\textsci{}b\textbari{}}\textprimstress{}\textrtails{}of]) (May 16 [O.S.\@ May 4] 1821 – December 8 [O.S.\@ November 26] 1894) was a Russian mathematician. His name can be alternatively transliterated as Chebychev, Chebysheff, Chebychov, Chebyshov; or Tchebychev, Tchebycheff (French transcriptions); or Tschebyschev, Tschebyschef, Tschebyscheff (German transcriptions).
\end{displayquote}

This is one example why characters which are not typical for one language can
appear in it. In this case, Cyrillic characters and special IPA symbols.

It is, however, expected that those characters appear less often than the ones
which form the language. Hence the set of most common characters~$C_{\theta}$
with coverage $\theta \in (0, 1]$ of a language can be formed as follows:
\begin{enumerate}
    \item Count all characters $c \in C$. The number of occurrences of $c$ is
          denoted by $n_c \in \mathbb{N}$.
    \item Sort the characters descending by $n_c$
    \item Define the minimum desired coverage as\\$\Sigma_{\theta} := \theta \cdot \left (\sum_{c \in C} n_c \right)$
    \item Initialize $C_{\theta}$ as an empty set and a counter $n = 0$
    \item Go through all characters. Add the character to $C_{\theta}$ if the
          $n < \Sigma_{\theta}$. Increase $n$ by $n_c$.
\end{enumerate}

For English, this leads to
\begin{itemize}
    \item $|C_{0.75}| = 13; C_{0.75} = \{$\texttt{SPACE}, a,c,d,e,h,i,l,n,o,r,s,t $\}$
    \item $|C_{0.80}| = 15; C_{0.80} = C_{0.75} \cup \{$m, u$\}$
    \item $|C_{0.90}| = 22; C_{0.90} = C_{0.80} \cup \{$\texttt{COMMA}, b, f, g, p, w,y$\}$
    \item $|C_{0.99}| = 58$; $C_{0.99} = C_{0.90} \cup \{$", ', (, ), -, ., 01234789, k, v, x$\}$ and ABCDEFGHIJLMNOPRSTW
    \item $|C_{1.00}| = 239$; including Arabic and Cyrillic letters and others.
\end{itemize}

Please note that $C_{0.99}$ still misses the characters \texttt{KQUVXYZjq}.
Especially the missing \texttt{Y} is most likely an effect of the source of the
data. I expect \texttt{Y} and \texttt{I} to appear much more often in personal
texts due to \texttt{You} and \texttt{I}.

$C_{0.99}$ is for most languages between \num{41}~characters (Lojban) and
150~characters.

There are only 9~languages in the dataset for which $C_{0.99}$ contains at
least 150~characters:
\begin{itemize}
    \setlength\itemsep{-0.2em}
    \item Amharic (280~characters), Ge'ez script
    \item Hakka Chinese (582~characters)
    \item Min Dong (762~characters)
    \item Korean (1158~characters)
    \item Japanese (1630~characters)
    \item Cantonese (2519~characters)
    \item Standard Chinese (2814~characters)
    \item Wu Chinese (3249~characters)
    \item Literary Chinese (3324~characters)
\end{itemize}

\begin{figure}
    \includegraphics[width=\linewidth]{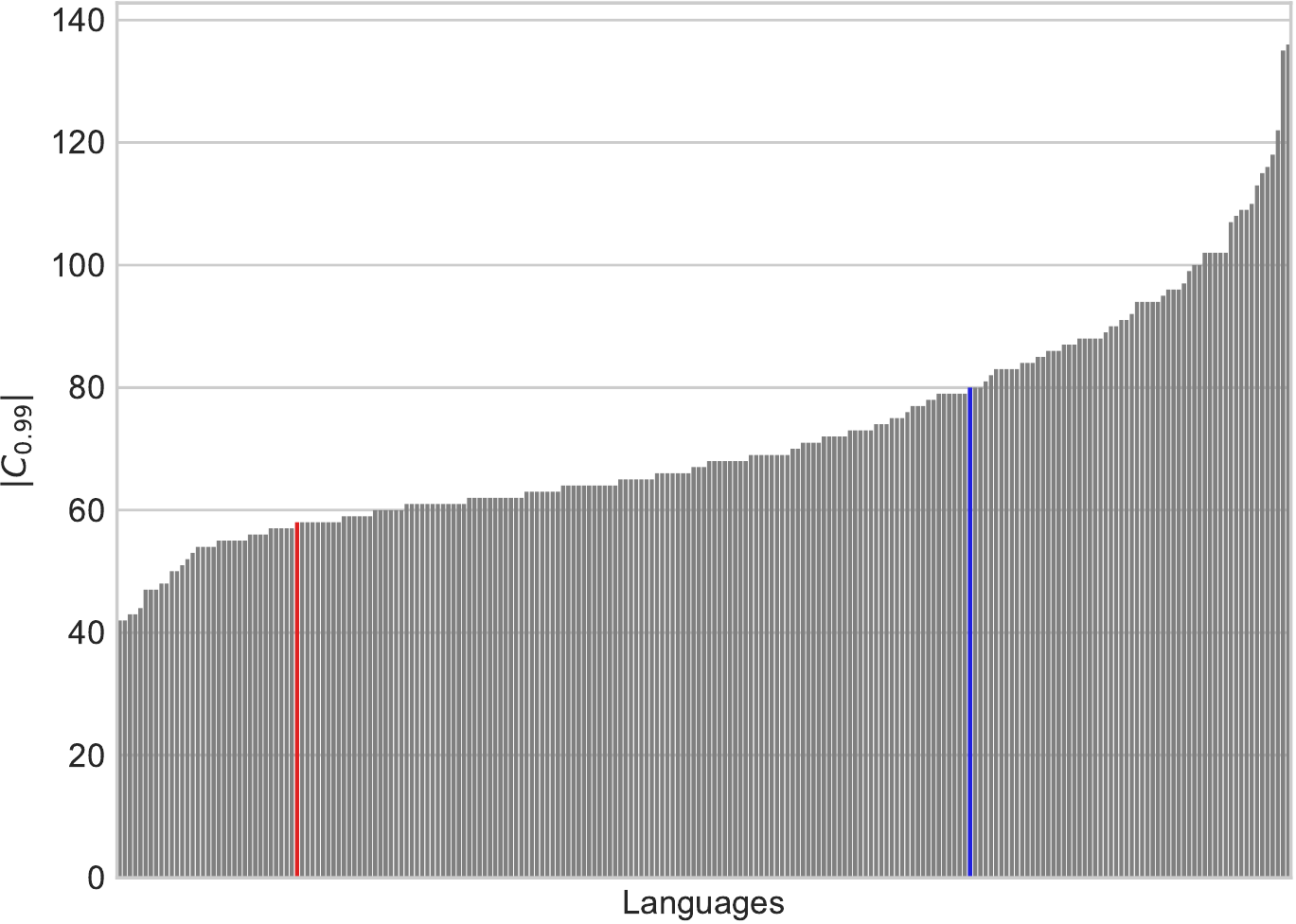}
    \caption{The distribution of characters per language. The red bar is
             English, the blue bar is Russian.}
    \label{fig:c-99-dist}
\end{figure}

\Cref{fig:c-99-dist} shows how many characters the other languages have.

The fact that those 9~languages have so much more different characters
suggests to treat those different.

\begin{figure}
    \includegraphics[width=\linewidth]{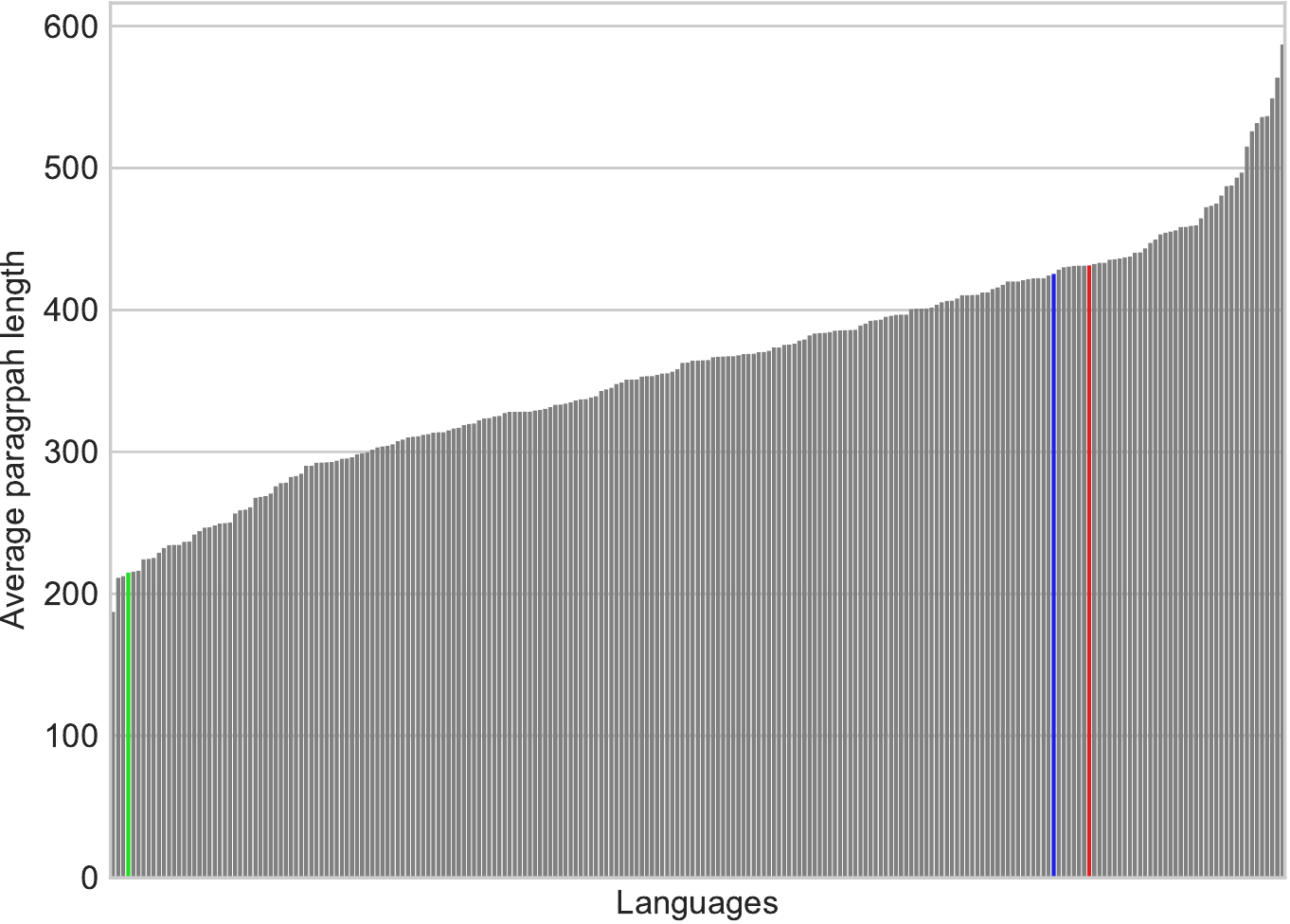}
    \caption{Mean paragraph length per language. The red bar is
             English, the blue bar is Russian and the green bar is standard
             Chinese. One can see that the average length of a paragraph
             is vastly different between languages. The following three languages
             are excluded from the graph: Pushto (621~characters),
             Uighur (750~characters) and Tibetan (2084~characters).}
    \label{fig:paragraph-length}
\end{figure}

The number of characters per paragraph shows that Tibetan has an average count
of characters per paragraph of \num{2084} which is far bigger than any other
language. Languages with many symbols are expected to have less characters per
paragraph. Another reason for the differences seen
in~\cref{fig:paragraph-length} could be writing styles and topics about which
was written.

\section{Errors and problems of the dataset}\label{sec:dataset-errors}
The following kinds of problems were observed in the dataset:
\begin{enumerate}[label=(E\arabic*)]
    \item Sources: Literature and sources are extremely hard to classify right.
          For example,
          \begin{addmargin}[1em]{2em}
          The Lycoming County Unit of the Pennsylvania Writers Project of the Work Projects Administration: A Picture of Lycoming County (= SSRI workshop series. Nr. 7512). 1. Auflage. The Commissioners of Lycoming County Pennsylvania, 1972, OCLC 17171801 (Erstausgabe: 1939).
          \end{addmargin}
          belongs to German because of \texttt{1. Auflage} and
          \texttt{Erstausgabe} which indicates that everything else are just
          names.\label{item:error-sources}
    \item Copy-paste errors: Wikipedia authors sometimes copy the content of \label{item:copy-paste}
          another language Wikipedia to their language. For example, in the
          Sundanese Wikipedia (\texttt{su}), there are several examples.\footnote{\url{https://su.wikipedia.org/w/index.php?title=Optimisasi_(matematik)&oldid=498807}, \url{https://su.wikipedia.org/w/index.php?title=Iron_Maiden&oldid=533732}}
    \item Translations: The Tongan Wikipedia has several pages which have an English translation.\footnote{Examples: \url{https://to.wikipedia.org/wiki/Sione_Tupou_Mateialona/en}, \url{https://to.wikipedia.org/wiki/Uho_taha/en}} Due to the large automatic labeling of sentences this means many English items would be wrongly labeled as Tonganese.
    \item Quotes: Wikipedia articles contain quotes. If the original quote is
          not in the language of the Wikipedia, then the original quote is
          often followed by a translated version. Due to the way how the
          paragraphs are extracted for \dbNameVersion{}, this leads to
          paragraphs being assigned to the wrong
          language.\label{item:error-quotes}
    \item Bias: People have different writing styles and different topics in
          which they are interested. For small Wikipedias, this can lead to a
          bias in the recognition of a language. For example, it is expected
          that articles about mathematics have different properties than
          articles about literature.
\end{enumerate}

\section{Written Language Identification Pipeline}
\begin{figure}
    \includegraphics[width=\linewidth]{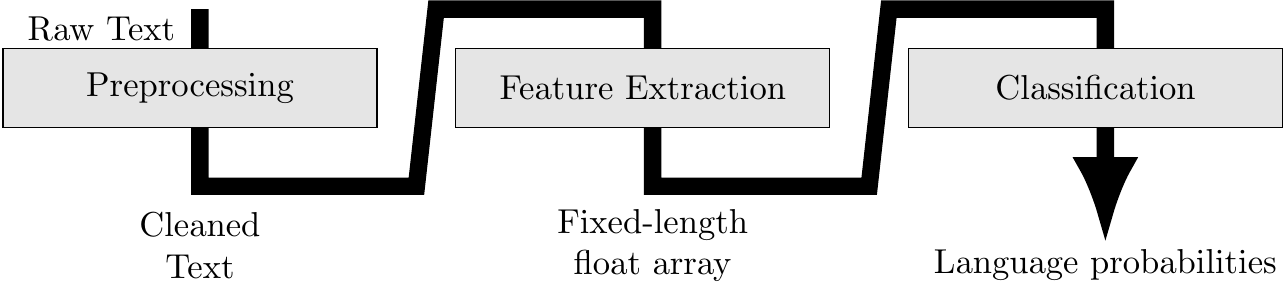}
    \caption{A typical Written Language Identification pipeline.}
    \label{fig:lid-architecture}
\end{figure}

A typical pipeline for written language identification first cleans the text
from unwanted characters and does normalization steps such as removing
punctuation, making the text lowercase or reducing the character set. The next
step is to create a vector from the filtered string. This can be as simple as
counting the number of occurrences of a word. Other relevant ideas are tf-idf
features and tokenizers. This vector can be classified by neural networks such
as multilayer Perceptrons (MLPs) or a Naïve Bayes classifier. This is
visualized in \cref{fig:lid-architecture}.

It is noteworthy that recurrent neural networks (RNNs) do not need a
fixed-length array as input.

\section{Evaluation}\label{sec:evaluation}
The code used for the evaluation can be found at~\cite{lidtk,clana}. The neural
networks were implemented with Keras~2.0.6~\cite{chollet2015keras} and
Tensorflow~1.2.0~\cite{tensorflow2015-whitepaper} as the backend. Other
algorithms such as the tf-idf vectorizer are implemented in
sklearn~0.19.1~\cite{scikit-learn}.

\subsection{Language Groups by Script}
A major difficulty in building a system for written natural language
identification is the size of Unicode~10.0. With \num{136755}~characters, the
feature space is large.

\Cref{table:unicode-blocks} shows for some Unicode blocks which languages
within \dbNameVersion{} use those blocks and to which extend.

\begin{table}
    \begin{tabular}{rrrp{3.2cm}l}
    \toprule
    Start & End  & \#         & High Coverage                                       & Next coverage \\\midrule
       0  &  879 & 880        & 150~languages ($\geq \SI{91}{\percent}$)            & $\leq \SI{48.20}{\percent}$\\
     880  & 1023 & 144        & Modern Greek ($\geq \SI{76}{\percent}$)             & $\leq \SI{0.52}{\percent}$\\
    1040  & 1103 &  64        & 31~languages with Russian Cyrillic script ($\geq \SI{60}{\percent}$) & $\leq \SI{5.36}{\percent}$\\
    1328  & 1423 &  96        & Armenian ($\geq \SI{78}{\percent}$)                 & $\leq \SI{0.06}{\percent}$\\
    1424  & 1535 & 112        & Yiddish and Hebrew ($\geq \SI{76}{\percent}$)       & $\leq \SI{0.17}{\percent}$\\
    1536  & 1791 & 256        & 13~languages with\newline{}Arabic script ($\geq \SI{69}{\percent}$) & $\leq \SI{0.55}{\percent}$\\
    1920  & 1983 &  64        & Dhivehi ($\geq \SI{86}{\percent}$)                  & $\leq \SI{0.06}{\percent}$\\
    1984  & 2431 & 448        & 9~languages ($\geq \SI{49}{\percent}$)              & $\leq \SI{1.50}{\percent}$\\
    2432  & 2559 & 128        & Bishnupriya, Bengali and Assamese ($\geq \SI{73}{\percent}$) & $\leq \SI{0.04}{\percent}$\\
    2560  & 2687 & 128        & Panjabi ($\geq \SI{74}{\percent}$)                  & $\leq \SI{0.04}{\percent}$\\
    2688  & 2815 & 128        & Gujarati ($\geq \SI{80}{\percent}$)                 & $\leq \SI{0.01}{\percent}$\\
    2816  & 2943 & 128        & Oriya ($\geq \SI{78}{\percent}$)                    & $\leq \SI{0.00}{\percent}$\\
    2944  & 3071 & 128        & Tamil ($\geq \SI{84}{\percent}$)                    & $\leq \SI{0.22}{\percent}$\\
    3072  & 3199 & 128        & Telugu ($\geq \SI{80}{\percent}$)                   & $\leq \SI{0.01}{\percent}$\\
    3200  & 3327 & 128        & Kannada and Tulu ($\geq \SI{83}{\percent}$)         & $\leq \SI{1.61}{\percent}$\\
    3328  & 3455 & 128        & Malayalam ($\geq \SI{84}{\percent}$)                & $\leq \SI{0.02}{\percent}$\\
    3456  & 3583 & 128        & Sinhala ($\geq \SI{76}{\percent}$)                  & $\leq \SI{0.00}{\percent}$\\
    3584  & 3711 & 128        & Thai ($\geq \SI{97}{\percent}$), Lao ($\geq \SI{7}{\percent}$) & $\leq \SI{0.05}{\percent}$\\
    3712  & 3839 & 128        & Lao ($\geq \SI{67}{\percent}$)                      & $\leq \SI{0.02}{\percent}$\\
    3840  & 4095 & 256        & Tibetan ($\geq \SI{97}{\percent}$)                  & $\leq \SI{0.02}{\percent}$\\
    12352 & 12543 &  192      & Japanese ($\geq \SI{49}{\percent}$)                 & $\leq \SI{0.10}{\percent}$\\
    19000 & 44000 & \num{12000} & Literary Chinese (\texttt{lzh}), Wu Chinese,
          Standard Chinese (\texttt{zho}) and Cantonese (\texttt{zh-yue}) ($\geq \SI{70}{\percent}$), Japanese ($\geq \SI{32}{\percent}$) & $\leq \SI{5.46}{\percent}$\\
    44000 & 56000 & \num{25000} & Korean ($\geq \SI{64}{\percent}$)                 & $\leq \SI{0.06}{\percent}$\\
    \bottomrule
    \end{tabular}
    \caption{Unicode blocks which differentiate languages well: The percentage
             in the \enquote{High Coverage} column denotes the amount of
             training data which is covered by characters in the Unicode block.
             The column \enquote{Next coverage} denotes the maximum training
             coverage any other language has in the given Unicode block.}
    \label{table:unicode-blocks}
\end{table}

This information can be used in several ways:
\begin{itemize}
    \item \textbf{Feature Engineering}: For a given range which only contains
          one language, all Unicode code points can be mapped to the first
          character of that range. This would reduce the feature space.
    \item \textbf{Hierarchical Classification}: A first classifier could try
          to identify the 15~languages which have their own Unicode blocks. For
          the 8~blocks which contain multiple languages one could make 8 other
          classifiers which have a very reduced feature space and much less
          classes to distinguish.
\end{itemize}

\subsection{Single-Character Frequency Analysis}

The analysis of the distribution of single characters is used for many years in
cryptanalysis and known as \textit{frequency analysis}~\cite{gaines2014cryptanalysis}.
The most simple approach to use frequency analysis for the identification of
the language of a text defines a set~$C$ of characters of
interest, counts them for a corpus for each language and compares the character
distribution for each language with the character distribution in the text of
an unknown language.

The two open choices in this approach are the choice of characters in~$C$ and
the choice of a distance function:
\[f: [0, 1]^{|D|} \times [0,1]^{|D|} \rightarrow \mathbb{R}\]
$f$ is applied to the character frequency distribution measured in the unknown
text and the character distribution of the language. It returns how close the
unknown texts character distribution is to the candidate language.

One way to define a character set is to count each character in the training
set by language. Then order the characters descending by the number of
occurrences in that language and take enough characters to reconstruct at least
$\theta \in (0, 1]$ of the languages training texts. A special \texttt{other}
character is used for all other characters. Thus $C$ is the union of all languages
$C_\theta$.

An intuitive choice for $f$ is the inverse discrete overlap
\[\ido(x, y) := 1 - \sum_{i=1}^n \min(x^{(i)}, y^{(i)})\]
which measures the overlap of the probability distributions $x$ and $y$. $\ido$ is
a metric~\cite{2409553} and almost equivalent to the L1 norm / city block distance
when $x$ and $y$ are discrete probability distributions:
\begin{align}
  \ido(x, y) &= 1 - \sum_{i=1}^n \min(x^{(i)}, y^{(i)})\\
             &= 1 - \sum_{i=1}^n \frac{x^{(i)} + y^{(i)} - |x^{(i)} - y^{(i)}|}{2}\\
             &= 1 - \frac{1}{2} \left (2 - \sum_{i=1}^n |x^{(i)} - y^{(i)}| \right )\\
             &= \frac{1}{2} \left\Vert x - y \right\Vert_1
\end{align}

The city block distance and 8~other metrics were compared against
each other in~\cref{table:char-measure-approach-results}.

\begin{table}[H]
    \begin{tabular}{@{}lrrrrr}
    \toprule
    \multirow{2}{*}{Metric} & Min   & \multirow{2}{*}{Characters} & \multirow{2}{*}{Accuracy} & Time per\\
                        & Coverage $\theta$ &            &                       & prediction \\\midrule
    Canberra distance   & \SI{80}{\percent} & \num{2069} &  \SI{7.81}{\percent} & \SI{16.05}{\milli\second} \\
    Cosine distance     & \SI{80}{\percent} & \num{2069} & \SI{76.25}{\percent} & \SI{15.88}{\milli\second} \\
    Chebyshev distance  & \SI{80}{\percent} & \num{2069} & \SI{56.12}{\percent} &\SI{31.97}{\milli\second} \\
    Correlation         & \SI{80}{\percent} & \num{2069} & \SI{76.23}{\percent} &\SI{19.25}{\milli\second} \\
    Squared Euclidean   & \SI{80}{\percent} & \num{2069} & \SI{76.27}{\percent} & \SI{8.67}{\milli\second} \\
    Braycurtis distance & \SI{80}{\percent} & \num{2069} & \textbf{\SI{78.96}{\percent}} & \SI{10.69}{\milli\second} \\
    City block distance  & \SI{80}{\percent} & \num{2069} & \textbf{\SI{78.96}{\percent}} & \textbf{\SI{9.05}{\milli\second}} \\
    City block distance  &\SI{100}{\percent} &\num{10366} & \SI{79.67}{\percent} & \SI{30.26}{\milli\second} \\
    City block distance  & \SI{75}{\percent} & \num{1659} & \SI{78.48}{\percent} & \SI{7.20}{\milli\second} \\
    City block distance  & \SI{50}{\percent} &  \num{593} & \SI{70.52}{\percent} & \SI{4.37}{\milli\second} \\
    \bottomrule
    \end{tabular}
    \caption{This table shows the accuracy and prediction speed for different
             choices of metrics and used characters. All metrics are used from
             SciPy~1.0.0~\cite{scipy}}
    \label{table:char-measure-approach-results}
\end{table}

\subsection{Character tf-idf Features and MLP Classifier}
Character-based term frequency-inverse document frequency (tf-idf) features
with a minimum absolute occurrence of 100~times in the training dataset are
used. The resulting \num{2218}~features are L2~normalized.

A multilayer Perceptron with one hidden layer of 512~neurons followed by the
ReLU activation function and an output layer of \dbTotalClasses{}~neurons
followed by softmax is used as a classifier. This network has
\num{1256683}~parameters. The cross-entropy loss function and the Adam
optimizer~\cite{kingma2014adam} are used. The model is trained for 20~epochs
with a mini-batch size of~32.

This model achieves a test set accuracy of \SI{88.30}{\percent} with
\SI{23}{\minute} of training. The complete test set was evaluated within
\SI{2.5}{\minute}.

\Cref{fig:cm-tfidf-nn} shows a confusion matrix which was optimized with
confusion matrix ordering (CMO) as introduced in~\cite{Thom2017}. It clearly
shows that many that were predicted to be English had a ground-truth which is
not English. A manual investigation shows that for many of those, the
classifier was either right or it is reasonable to predict English due
to~\cref{item:error-sources}. A similar, but not as strong pattern can be seen
for Russian.

\section{Comparison of Available Solutions}
Ready to use tools for language identification support different sets of
languages. Simply taking the accuracy could draw a skewed picture of
the quality of the software. For this reason, every classifier is described by
the class-wise precision, recall, $F_1$-Score and the overall accuracy.
The class-wise precision is
$$\text{precision(c)} = \frac{\text{predicted class } c \text{ and is class } c}{\text{predicted class } c}\text{,}$$
the class-wise recall is defined as
$$\text{recall(c)} = \frac{\text{predicted class } c \text{ and is class } c}{\text{is class } c}\text{.}\label{eq:recall}$$

A high precision for class~$c$ means that one can trust the prediction of the
classifier if $c$ was predicted. A high recall for class~$c$ means one can
rely on the classifier to find examples of class~$c$. But a classifier could
simply always predict class~$c$, so only recall is meaningless. Hence the $F_1$
score is used to combine both. The $F_1$-score is the harmonic mean of both,
precision and recall:
$$F_1 = 2 \cdot \frac{\mathrm{precision} \cdot \mathrm{recall}}{\mathrm{precision} + \mathrm{recall}}$$

Languages which are more influential like English, Russian or French are
expected to have a lower precision on this dataset as there are red herrings
such as~\cref{item:error-quotes,item:copy-paste,item:error-sources}. Also
languages like German which have a lot of close languages such as its dialects
are expected to have a lower precision.

\begin{table}
\begin{tabular}{lccp{0.5cm}c}
\toprule
Name                                  & TextCat            & CLD-2               & lang\-detect       & langid\\\midrule
Languages in \dbNameVersion{}         & \textbf{140}       &  90                 & 55                 & 95\\
Languages not in \dbNameVersion{}     & \textbf{115}       &  10                 & \mbox{\hphantom{0}0} & \hphantom{0}2\\
Speed for $10^6$ elements             & $1.5 \cdot 10^6$s  &  \textbf{\SI{38}{\second}}   & \SI{8025}{\second} & \SI{1858}{\second}\\
\dag Mean Precision                   & \SI{48}{\percent}  &  \SI{96}{\percent}  & \textbf{\SI{97}{\percent}}  & \SI{93}{\percent}\\
\dag Mean Recall                      & \SI{42}{\percent}  &  \SI{95}{\percent}  & \textbf{\SI{96}{\percent}}  & \SI{90}{\percent}\\
\dag Mean F1                          & \SI{38}{\percent}  &  \SI{95}{\percent}  & \textbf{\SI{96}{\percent}}  & \SI{90}{\percent}\\
\dbNameVersion{} accuracy             & \SI{35}{\percent}  &  \textbf{\SI{36}{\percent}}  & \SI{22}{\percent}  & \textbf{\SI{36}{\percent}}\\
Unknown prediction accuracy           & ---                &  \textbf{\SI{37}{\percent}}  & \SI{25}{\percent}  & \textbf{\SI{37}{\percent}}\\\bottomrule
\end{tabular}
\caption{Comparison of the CLD-2, TextCat, langdetect and langid tools for
         language identification. The row \enquote{Unknown prediction accuracy}
         denotes the classifiers accuracy if all languages not know by
         the classifier expects unknown to be counted as correct.\\
         \dag: The \dbNameVersion{} test dataset was reduced to the set of
         languages supported by the classifier.}
\label{table:cld2-textcat-comparison}
\end{table}

\subsection{Textcat}
TextCat is the name of the software which implements the $n$-gram approach
described in~\cite{cavnar1994n}. The implementation of
NLTK~3.2.5~\cite{Pekker2001,steven2009natural} was used.

The TextCat implementation works as follows.

First, create language fingerprints:
\begin{itemize}
    \item Remove punctuation
    \item The text was tokenized into word-tokens
    \item Trigrams are counted
\end{itemize}

Then, apply the same process for a new text. Compare each of the language
fingerprints with the text of unknown language fingerprint. Use a rank-order
statistic called \textit{out-of-place} measure. The language with the least
distance is predicted.

There are four drawbacks of this approach:
\begin{itemize}
    \item It is unable to predict that it doesn't know a language.
    \item It does not consider the script of the text.
    \item The out-of-place measure is not well-suited to the problem as the
          less-common $n$-grams will have an arbitrary out-of-place measure.
    \item While some $n$-grams are rare, they could at the same time
          be a strong indicator for the language. For example, \texttt{über} is a
          strong indicator for German as it is the German word \texttt{over}
          and contains an umlaut. Hence tf-idf features are better suited.
\end{itemize}

\subsection{Compact Language Detector 2}
\href{https://github.com/CLD2Owners/cld2}{Compact Language Detector 2} (CLD2)
is an Apache 2.0 licensed C++ project.

CLD~2 uses a Naïve Bayes classifier and one of three different token algorithms:
\begin{itemize}
    \item Unicode scripts such as Greek and Thai that map one-to-one to
          detected languages: the script defines the result
    \item \num{80000}+ character Han script and its CJK combination with
          Hiragana, Katakana, and Hangul scripts: single letters (unigrams) are
          scored
    \item other scripts: sequences of four letters are scored
\end{itemize}

Preprocessing:
\begin{itemize}
    \item lowercased Unicode letters and marks
    \item deleting digits, punctuation
\end{itemize}

According to the Python binding provided by Michael McCandless and Greg Bowyer,
CLD-2 supports 282~languages. The Python binding, however, only predicted
100~languages at least once for \dbNameVersion{}.

CLD-2 confuses Serbian, Bosnian and Croatian. Another difficult group of
languages is Malay and Indonesian for CLD-2. Swedish is confused 28 out of 500
times with Bokmål and CLD-2 classified Danish as Bokmål 30 out of 500~times.

\subsection{langdetect}
\texttt{langdetect} is a Python wrapper created by Michal Danilak for the Java
library \texttt{language-detection}~\cite{Nakatani2015}. Both,
\texttt{langdetect} and the \texttt{language-detection} library is licensed
under the Apache License, Version~2.0 and was created in~2010 by Cybozu Labs,
Inc. \texttt{language-detection} supports 55~languages. It uses character
$n$-gram features and a Naïve Bayes classifier~\cite{Nakatani2010}. They added a
noise filter and character normalization.

For \texttt{langdetect}, the hardest group of classes is Korean, Literary
Chinese and Standard Chinese. For Standard Chinese, it has a recall of only
\SI{51}{\percent}.

Modern Standard Urdu has a recall of only \SI{89}{\percent} as it is often
confused to be Arabic.

In the reduced dataset that contained only the 55~languages supported by
langdetect, only two samples were predicted to be unknown languages.

\subsection{Online Services}
There are multiple online services for language identification.

The \href{https://open.xerox.com/Services/LanguageIdentifier}{Xerox Language Identifier Web Service}
is able to detect 80~languages (see \cpageref{sec:xerox-lid-web-service-languages}).

\href{https://detectlanguage.com}{detectlanguage.com} is a web service which
allows its users to detect the language of a text snipped via a POST request.
The service knows 164~languages, of which the following 42~languages are not
within \dbNameVersion{} (see~\cpageref{sec:dl-unknown-to-wili}) and
\dbNameVersion{} contains 89~languages which are not within detectlanguage.

Detectlanguage uses CLD2 as one component for language detection.

The Google Cloud Translation API supports 104~languages.

\subsection{langid.py}
Langid.py is a Python project under BSD license. It supports 97~languages which
it trained with data from JRC-Acquis, ClueWeb~09, Wikipedia, Reuters~RCV2 and
Debian~i18n~\cite{Lui2017,lui2012langid}.

Langid.py uses LD feature selection~\cite{lui2011cross} and a Naïve Bayes
classifier.

\section{Conclusion and Future Work}
\dbNameVersion{} can be used to evaluate systems for written language identification
of 235~languages.\footnote{English, however, should be ignored due to
\cref{item:error-quotes,item:copy-paste,item:error-sources}} CLD-2 is by far
the fastest LID system and supports 100~languages. If support for more or other
languages is needed, then \texttt{lidtk}~\cite{lidtk} in combination with
\dbNameVersion{} is a starting point.

Future work with \dbNameVersion{} includes the evaluation of RNNs and Markov
Models, building a classifier which can reliably distinguish known from unknown
languages and examining the influence of text length and number of samples on
the efficacy of different classifiers. The last two points are extremely
relevant for minority languages.

\section{Acknowledgment}
I want to thank \enquote{Begabtenstiftung Informatik Karls\-ruhe}, the
Foundation for Gifted Informatics Students in Karlsruhe. Their support helped
me to write this work.

\bibliographystyle{IEEEtranSA}
\bibliography{literatur}

\appendix
\section*{Obtaining the data}\label{sec:obtain-dataset}
The data can be found at \dbDownloadURL{}. It is a \verb+tar.gz+ file of
\SI{\dbSizeMB}{\mega\byte}. The file can be verified with the MD5sum

\texttt{\dbMDfivesum}

The data is published under the ODbL~license. If you use
the \dbName~dataset, please cite this paper.

The \verb+zip+ archive contains all data as four UTF-8 encoded CSV files
(\texttt{x\_train.csv}, \texttt{y\_train.csv}, \texttt{x\_test.csv}, \texttt{y\_test.csv}).
Each line of the \texttt{x\_train.csv} / \texttt{x\_test.csv} files contains
one paragraph with at least 140 Unicode code points written in one language.
That laguage is provided in \texttt{y\_train.csv} / \texttt{y\_test.csv} as
ISO 369-3 code if available, otherwise as the Wikipedia code.

\section*{Languages}\label{sec:languages}
The following \dbTotalClasses{}~languages are part of the \dbName{}~dataset:

Achinese, Afrikaans, Albanian, Amharic, Arabic, Aragonese, Armenian, Aromanian,
Arpitan, Assamese, Asturian, Avar, Aymara, Azerbaijani, Banjar, Banyumasan,
Bashkir, Basque, Bavarian, Belarusian, Belarusian (Taraschkewiza), Bengali,
Bhojpuri, Bishnupriya, Bokmål, Bosnian, Breton, Bulgarian, Burmese, Buryat,
Cantonese, Catalan, Cebuano, Central Bikol, Central Khmer, Central Kurdish,
Chavacano, Chechen, Cherokee, Chinese, Chuvash, Classical Nahuatl, Cornish,
Corsican, Crimean Tatar, Croatian, Czech, Danish, Dhivehi, Dimli, Doteli,
Dutch, Eastern Mari, Egyptian Arabic, Emilian, English, Erzya, Esperanto,
Estonian, Extremaduran, Faroese, Fiji Hindi, Finnish, French, Friulian, Gagauz,
Galician, Georgian, German, Gilaki, Guarani, Gujarati, Haitian Creole, Hakka
Chinese, Hausa, Hebrew, Hindi, Hungarian, Icelandic, Ido, Igbo, Iloko,
Indonesian, Interlingua, Interlingue, Irish, Italian, Jamaican Patois,
Japanese, Javanese, Kabardian, Kabyle, Kannada, Karachay-Balkar, Karakalpak,
Kashubian, Kazakh, Kinyarwanda, Kirghiz, Komi, Komi-Permyak, Konkani, Korean,
Kurdish, Ladino, Lao, Latgalian, Latin, Latvian, Lezghian, Ligurian, Limburgan,
Lingala, Literary Chinese, Lithuanian, Livvi-Karelian, Lojban, Lombard, Low
German, Lower Sorbian, Luganda, Luxembourgish, Macedonian, Maithili, Malagasy,
Malay, Malayalam, Maltese, Manx, Maori, Marathi, Mazanderani, Min Dong, Min Nan
Chinese, Minangkabau, Mingrelian, Mirandese, Modern Greek, Moksha, Mongolian,
Narom, Navajo, Neapolitan, Nepali (macrolanguage), Newari, Northern Luri,
Northern Sami, Northern Sotho, Norwegian Nynorsk, Occitan, Old English , Oriya,
Oromo, Ossetian, Palatine German, Pampanga, Pangasinan, Panjabi, Papiamento,
Pennsylvania German, Persian, Picard, Polish, Portuguese, Pushto, Quechua,
Ripuarisch, Romanian, Romansh, Russian, Rusyn, Samogitian, Sanskrit, Sardinian,
Saterfriesisch, Scots, Scottish Gaelic, Serbian, Serbo-Croatian, Shona,
Sicilian, Silesian, Sindhi, Sinhala, Slovak, Slovene, Somali, South
Azerbaijani, Spanish, Sranan, Sundanese, Swahili (macrolanguage), Swedish,
Tagalog, Tajik, Tamil, Tarantino dialect, Tatar, Telugu, Tetum, Thai, Tibetan,
Tongan, Tosk Albanian, Tswana, Tulu, Turkish, Turkmen, Tuvan, Udmurt, Uighur,
Ukrainian, Upper Sorbian, Urdu, Uzbek, Venetian, Veps, Vietnamese, Vlaams,
Volapük, Võro, Walloon, Waray, Welsh, West Low German, Western Frisian, Western
Mari, Western Panjabi, Wolof, Wu Chinese, Xhosa, Yakut, Yiddish, Yoruba,
Zeeuws.

\subsection*{Detectlanguage Web Service}\label{sec:dl-unknown-to-wili}
The folowing 42~languages are not within the \dbName{}~dataset, but within
detectlanguage.com:

Afar, Abkhazian, Akan, Bislama, Buginese, Cherokee, Seselwa, Dzongkha,
Gothic, Hausa, Hawaiian, Hmong, Igbo, Inupiak, Inuktitut,
Khasi, Greenlandic, Kashmiri, Ganda, Limbu, Mauritian Creole, Nauru, Ndebele,
Nyanja, Oromo, Rundi, Kinyarwanda, Sango, Samoan, Siswant, Sesotho, Syriac,
Tigrinya, Klingon, Tswana, Tonga, Tsonga, Venda, Wolof, Xhosa, Zhuang, Zulu

\subsection*{Xerox Language Identificatin Web Service}\label{sec:xerox-lid-web-service-languages}
The 80~languages of the Xerox Language Identificatin Web Service:

Afrikaans, Albanian, Arabic, Armenian, Azerbaijani, Basque, Belarusian,
Bosnian, Breton, Bulgarian, Burmese, Catalan, Cebuano, Chinese, Croatian,
Czech, Danish, Dutch, English, Esperanto, Estonian, Finnish, French, Galician,
Georgian, German, Greek, Haitian, Hebrew, Hindi, Hungarian, Icelandic,
Indonesian, Irish, Italian, Japanese, Javanese, Kazakh, Korean, Latin, Latvian,
Lithuanian, Luxembourgish, Macedonian, Malagasy, Malay, Malayalam, Maltese,
Marathi, Minangkabau, Nepal Bhasa (Newari), Norwegian, Norwegian Nynorsk,
Occitan, Persian, Polish, Portuguese, Quechua, Romanian, Russian, Serbian,
Slovak, Slovenian, Spanish (Castilian), Swahili, Swedish, Tagalog, Tajik,
Tamil, Tatar, Telugu, Thai, Turkish, Ukrainian, Urdu, Uzbek, Vietnamese, Waray,
Welsh, Yorubac

\subsection*{CLD2}\label{sec:CLD2-languages}
Afrikaans, Albanian, Amharic, Arabic, Armenian, Azerbaijani, Basque,
Belarusian, Bengali, Bhojpuri, Bokmål, Bosnian, Bulgarian, Burmese, Catalan,
Cebuano, Central Khmer, Cherokee, Croatian, Czech, Danish, Dhivehi, Dutch,
English, Estonian, Finnish, French, Galician, Georgian, German, Gujarati,
Haitian Creole, Hebrew, Hindi, Hungarian, Icelandic, Indonesian, Irish,
Italian, Japanese, Javanese, Kannada, Kazakh, Kinyarwanda, Kirghiz, Korean,
Kurdish, Lao, Latvian, Literary Chinese, Lithuanian, Luganda, Macedonian,
Malagasy, Malay, Malayalam, Maltese, Marathi, Modern Greek, Nepali
(macrolanguage), Oriya, Panjabi, Persian, Polish, Portuguese, Romanian,
Russian, Scottish Gaelic, Serbian, Sinhala, Slovak, Slovene, Spanish, Standard
Chinese, Sundanese, Swahili (macrolanguage), Swedish, Tagalog, Tajik, Tamil,
Telugu, Thai, Tibetan, Turkish, Ukrainian, Urdu, Uzbek, Vietnamese, Welsh,
Yiddish

\subsection*{Langdetect}\label{sec:langdetect-languages}
Afrikaans, Albanian, Arabic, Bengali, Bokmål, Bulgarian, Catalan, Croatian,
Czech, Danish, Dutch, English, Estonian, Finnish, French, German, Gujarati,
Hebrew, Hindi, Hungarian, Indonesian, Italian, Japanese, Kannada, Korean,
Latvian, Literary Chinese, Lithuanian, Macedonian, Malayalam, Marathi, Modern
Greek, Nepali (macrolanguage), Panjabi, Persian, Polish, Portuguese, Romanian,
Russian, Slovak, Slovene, Somali, Spanish, Standard Chinese, Swahili
(macrolanguage), Swedish, Tagalog, Tamil, Telugu, Thai, Turkish, Ukrainian,
Urdu, Vietnamese, Welsh

\subsection*{Textcat}\label{sec:textcat-languages}
Languages within WiLI:

Achinese, Afrikaans, Alemannic German, Amharic, Aragonese, Armenian, Aromanian,
Arpitan, Assamese, Asturian, Avar, Bashkir, Basque, Bavarian, Belarusian,
Bengali, Bokmål, Bosnian, Breton, Bulgarian, Burmese, Buryat, Catalan, Cebuano,
Central Bikol, Central Khmer, Chuvash, Cornish, Corsican, Croatian, Czech,
Danish, Dimli, Dutch, English, Esperanto, Estonian, Faroese, Finnish, French,
Friulian, Gagauz, Galician, Georgian, German, Gujarati, Haitian Creole, Hausa,
Hebrew, Hindi, Hungarian, Icelandic, Igbo, Iloko, Indonesian, Interlingua,
Irish, Italian, Javanese, Kabardian, Kabyle, Kannada, Karachay-Balkar,
Kashubian, Kazakh, Kinyarwanda, Kirghiz, Ladino, Lao, Latin, Latvian, Lingala,
Lithuanian, Lombard, Luganda, Luxembourgish, Macedonian, Malayalam, Maltese,
Manx, Maori, Minangkabau, Modern Greek, Navajo, Neapolitan, Northern Sami,
Northern Sotho, Norwegian Nynorsk, Old English , Oriya, Ossetian, Pampanga,
Panjabi, Papiamento, Pennsylvania German, Polish, Portuguese, Ripuarisch,
Romanian, Romansh, Russian, Scots, Scottish Gaelic, Serbian, Shona, Sicilian,
Sindhi, Slovak, Slovene, Somali, Spanish, Sundanese, Swedish, Tagalog, Tajik,
Tamil, Tatar, Telugu, Tetum, Thai, Tibetan, Tongan, Tswana, Turkish, Turkmen,
Udmurt, Uighur, Ukrainian, Upper Sorbian, Urdu, Venetian, Vietnamese, Vlaams,
Welsh, West Low German, Western Mari, Wolof, Xhosa, Yoruba, Zeeuws

ISO 369-3 codes of languages not in WiLI: abn, ach, agr, aka, als, ami, amr,
arb, ayr, azj, bam, ban, bba, bcc, bis, buc, byv, cbr, cha, cmn, cnh, crs, csa,
ddn, dhv, dyu, dzo, emk, eml, ewe, fij, fri, frr, fub, fud, gaz, gjn, gkn, got,
gsc, gug, guw, haw, hil, hmo, hne, ivv, kal, khk, kjh, kmr, knc, knn, kpv, kri,
lld, lms, lnc, loz, lua, lus, mcd, mfe, mho, mic, mly, mus, naq, nbl, nde, nia,
nmf, nya, nyk, ojw, ood, pau, pbb, pbu, pcm, pes, pih, pis, plt, pms, pon, prq,
prs, prv, quh, rar, rnd, sot, src, srm, srr, ssw, sum, sus, swb, swh, teo, tig,
tos, tsc, tso, tvl, tzc, tzm, umb, uzn, ven, yaf, ydd, zul

\section*{Long results}
\subsection*{CLD-2}
\begin{table}
    \begin{tabular}{llll|llll}
    \toprule
    \textbf{Lang}  & \textbf{Prec} & \textbf{Recall} & $\mathbf{F}_1$ & \textbf{Lang}  & \textbf{Prec} & \textbf{Recall} & $\mathbf{F_1}$ \\\midrule
    afr  & 0.65 & 0.96 & 0.78 & kor & 0.93 & 0.98 & 0.95 \\
    ara  & 0.41 & \textcolor{ForestGreen}{1.00} & 0.58 & kur & 0.68 & 0.97 & 0.80 \\
    aze  & 0.70 & 0.98 & 0.82 & lao & \textcolor{ForestGreen}{1.00} & 0.78 & 0.88 \\
    bel  & 0.47 & 0.92 & 0.62 & lav & 0.91 & 0.97 & 0.94 \\
    ben  & 0.39 & 0.90 & 0.54 & lit & 0.75 & 0.95 & 0.84 \\
    bho  & 0.67 & 0.59 & 0.63 & mal & 0.99 & 0.98 & 0.98 \\
    bos  & 0.38 & \textcolor{red}{0.41} & 0.39 & mar & 0.60 & 0.97 & 0.74 \\
    bul  & 0.81 & 0.92 & 0.86 & mkd & 0.96 & 0.95 & 0.95 \\
    cat  & 0.54 & 0.94 & 0.69 & mlg & 0.96 & 0.99 & 0.98 \\
    ceb  & 0.66 & \textcolor{red}{0.48} & 0.56 & mlt & 0.97 & 0.98 & 0.98 \\
    ces  & 0.95 & 0.87 & 0.91 & msa & 0.66 & 0.88 & 0.76 \\
    cym  & 0.94 & 0.95 & 0.95 & mya & 0.99 & 0.99 & \textcolor{ForestGreen}{0.99} \\
    dan  & 0.83 & 0.92 & 0.87 & nep & 0.36 & 0.97 & 0.53 \\
    deu  & \textcolor{red}{0.17} & 0.97 & 0.29 & nld & 0.25 & 0.97 & 0.39 \\
    div  & \textcolor{ForestGreen}{1.00} & 0.98 & 0.99 & nob & 0.38 & 0.96 & 0.55 \\
    ell  & 0.60 & 0.95 & 0.74 & ori & \textcolor{ForestGreen}{1.00} & 0.98 & 0.99 \\
    eng  & \textcolor{red}{0.07} & 0.98 & \textcolor{red}{0.14} & pan & 0.99 & 0.97 & 0.98 \\
    est  & 0.56 & 0.95 & 0.71 & pol & 0.74 & 0.96 & 0.84 \\
    eus  & 0.97 & 0.99 & 0.98 & por & 0.62 & 0.96 & 0.76 \\
    fas  & 0.20 & 0.99 & 0.34 & ron & 0.61 & 0.93 & 0.74 \\
    fin  & 0.59 & 0.99 & 0.74 & rus & 0.23 & 0.98 & 0.37 \\
    fra  & 0.22 & 0.97 & 0.36 & sin & \textcolor{ForestGreen}{1.00} & 0.97 & 0.99 \\
    gla  & 0.97 & 0.97 & 0.97 & slk & 0.97 & 0.97 & 0.97 \\
    gle  & 0.95 & 0.98 & 0.96 & slv & 0.99 & 0.92 & 0.95 \\
    glg  & 0.47 & 0.95 & 0.63 & spa & 0.17 & 0.92 & \textcolor{red}{0.29} \\
    guj  & 0.99 & 0.99 & 0.99 & sqi & 0.98 & 0.97 & 0.97 \\
    hat  & 0.97 & 0.97 & 0.97 & srp & 0.63 & 0.93 & 0.75 \\
    heb  & 0.99 & 0.99 & \textcolor{ForestGreen}{0.99} & sun & 0.77 & 0.87 & 0.82 \\
    hin  & 0.58 & \textcolor{ForestGreen}{1.00} & 0.73 & swa & 0.71 & 0.88 & 0.79 \\
    hrv  & 0.40 & 0.63 & 0.49 & swe & 0.93 & 0.89 & 0.91 \\
    hun  & 0.98 & 0.98 & 0.98 & tam & 0.98 & 0.98 & 0.98 \\
    hye  & 0.98 & 0.99 & 0.98 & tel & 0.99 & 0.97 & 0.98 \\
    ind  & 0.27 & 0.84 & 0.41 & tgk & 0.99 & 0.95 & 0.97 \\
    isl  & 0.54 & 0.99 & 0.70 & tgl & 0.42 & 0.97 & 0.59 \\
    ita  & \textcolor{red}{0.13} & 0.93 & \textcolor{red}{0.22} & tha & 0.92 & \textcolor{ForestGreen}{1.00} & 0.96 \\
    jav  & 0.60 & 0.95 & 0.73 & tur & 0.30 & 0.99 & 0.46 \\
    jpn  & 0.76 & 0.99 & 0.86 & ukr & 0.57 & 0.96 & 0.71 \\
    kan  & 0.50 & \textcolor{ForestGreen}{1.00} & 0.66 & urd & 0.40 & 0.94 & 0.57 \\
    kat  & 0.49 & 0.98 & 0.65 & uzb & 0.78 & 0.99 & 0.87 \\
    kaz  & 0.75 & 0.99 & 0.85 & vie & 0.95 & 0.97 & 0.96 \\
    khm  & \textcolor{ForestGreen}{1.00} & 0.79 & 0.88 & yid & \textcolor{ForestGreen}{1.00} & \textcolor{ForestGreen}{1.00} & \textcolor{ForestGreen}{1.00} \\
    kir  & 0.44 & 0.99 & 0.61 & zho & 0.40 & \textcolor{red}{0.46} & 0.43 \\\hline
    ~    & ~    & ~    & ~    & ~   & ~    & ~    & ~    \\
    \bottomrule
    \end{tabular}
    \caption{Results of CLD-2 applied to \dbNameVersion.}
    \label{table:cld2-results}
\end{table}

\subsection*{Char-distribution}
\begin{table}
    \begin{tabular}{llll|llll}
    \toprule
    \textbf{Lang}  & \textbf{Prec} & \textbf{Recall} & $\mathbf{F}_1$ & \textbf{Lang}  & \textbf{Prec} & \textbf{Recall} & $\mathbf{F_1}$ \\\midrule
    ace       & 0.96      & 0.86   & 0.91     & lmo      & 0.76      & 0.65   & 0.70     \\
    afr       & 0.71      & 0.77   & 0.74     & lrc      & 0.97      & 0.53   & 0.69     \\
    als       & 0.70      & 0.68   & 0.69     & ltg      & 0.89      & 0.86   & 0.87     \\
    amh       & \textcolor{ForestGreen}{1.00} & 0.99   & 0.99     & ltz      & 0.75      & 0.57   & 0.65     \\
    ang       & 0.95      & 0.85   & 0.90     & lug      & 0.91      & 0.85   & 0.88     \\
    ara       & 0.78      & 0.94   & 0.85     & lzh      & 0.72      & 0.97   & 0.82     \\
    arg       & 0.58      & 0.55   & 0.57     & mai      & 0.65      & 0.79   & 0.71     \\
    arz       & 0.89      & 0.77   & 0.83     & mal      & \textcolor{ForestGreen}{1.00} & 0.98   & 0.99     \\
    asm       & \textcolor{ForestGreen}{1.00} & 0.92   & 0.96     & map-bms  & 0.45      & 0.40   & 0.43     \\
    ast       & 0.61      & 0.36   & 0.45     & mar      & 0.60      & 0.93   & 0.73     \\
    ava       & 0.27      & \textcolor{red}{0.04}   & \textcolor{red}{0.08} & mdf      & 0.81      & 0.82   & 0.81     \\
    aym       & 0.78      & 0.81   & 0.79     & mhr      & 0.63      & 0.76   & 0.69     \\
    azb       & 0.92      & 0.91   & 0.92     & min      & 0.52      & 0.81   & 0.64     \\
    aze       & 0.99      & 0.97   & 0.98     & mkd      & 0.53      & 0.95   & 0.68     \\
    bak       & 0.99      & 0.93   & 0.96     & mlg      & 0.99      & 0.99   & 0.99     \\
    bar       & 0.59      & 0.63   & 0.61     & mlt      & 0.97      & 0.96   & 0.96     \\
    bcl       & 0.61      & 0.69   & 0.64     & mon      & 0.96      & 0.94   & 0.95     \\
    be-tarask & 0.58      & 0.88   & 0.70     & mri      & 0.99      & 0.90   & 0.95     \\
    bel       & 0.76      & 0.30   & 0.43     & mrj      & \textcolor{ForestGreen}{1.00} & 0.89   & 0.94     \\
    ben       & 0.90      & 0.89   & 0.90     & msa      & 0.47      & 0.55   & 0.51     \\
    bho       & 0.86      & 0.25   & 0.39     & mwl      & 0.61      & 0.77   & 0.68     \\
    bjn       & 0.58      & 0.79   & 0.67     & mya      & \textcolor{ForestGreen}{1.00} & 0.98   & 0.99     \\
    bod       & \textcolor{ForestGreen}{1.00} & \textcolor{ForestGreen}{1.00} & \textcolor{ForestGreen}{1.00} & myv      & 0.82      & 0.74   & 0.78     \\
    bos       & 0.32      & 0.28   & 0.30     & mzn      & 0.56      & 0.80   & 0.66     \\
    bpy       & 0.98      & 0.91   & 0.94     & nan      & 0.95      & 0.86   & 0.91     \\
    bre       & 0.83      & 0.86   & 0.84     & nap      & 0.87      & 0.62   & 0.73     \\
    bul       & 0.93      & 0.82   & 0.87     & nav      & \textcolor{ForestGreen}{1.00} & \textcolor{ForestGreen}{1.00} & \textcolor{ForestGreen}{1.00} \\
    bxr       & 0.98      & 0.89   & 0.93     & nci      & 0.99      & 0.81   & 0.89     \\
    cat       & 0.63      & 0.57   & 0.60     & nds      & 0.79      & 0.72   & 0.75     \\
    cbk       & 0.30      & 0.27   & 0.29     & nds-nl   & 0.43      & 0.70   & 0.53     \\
    cdo       & \textcolor{ForestGreen}{1.00} & 0.91   & 0.95     & nep      & 0.60      & 0.55   & 0.58     \\
    ceb       & 0.68      & 0.56   & 0.61     & new      & 0.98      & 0.87   & 0.92     \\
    ces       & 0.81      & 0.75   & 0.78     & nld      & 0.50      & 0.64   & 0.56     \\
    che       & 0.91      & 0.94   & 0.92     & nno      & 0.52      & 0.64   & 0.58     \\
    chr       & \textcolor{ForestGreen}{1.00} & 0.97   & 0.98     & nob      & 0.44      & 0.45   & 0.45     \\
    chv       & 0.99      & 0.95   & 0.97     & nrm      & 0.80      & 0.72   & 0.76     \\
    ckb       & \textcolor{ForestGreen}{1.00} & 0.94   & 0.97     & nso      & 0.98      & 0.73   & 0.84     \\
    cor       & 0.96      & 0.87   & 0.91     & oci      & 0.68      & 0.57   & 0.62     \\
    cos       & 0.71      & 0.74   & 0.72     & olo      & 0.89      & 0.83   & 0.86     \\
    crh       & 0.92      & 0.56   & 0.70     & ori      & \textcolor{ForestGreen}{1.00} & 0.98   & 0.99     \\
    csb       & \textcolor{ForestGreen}{1.00} & 0.95   & 0.97     & orm      & 0.89      & 0.80   & 0.85     \\
    cym       & 0.98      & 0.96   & 0.97     & oss      & \textcolor{ForestGreen}{1.00} & 0.95   & 0.97     \\
    dan       & 0.63      & 0.68   & 0.65     & pag      & 0.73      & 0.63   & 0.67     \\
    deu       & 0.53      & 0.74   & 0.62     & pam      & 0.71      & 0.64   & 0.68     \\
    diq       & 0.98      & 0.84   & 0.90     & pan      & \textcolor{ForestGreen}{1.00} & 0.98   & 0.99     \\
    div       & \textcolor{ForestGreen}{1.00} & 0.98   & 0.99     & pap      & 0.68      & 0.73   & 0.71     \\
    dsb       & 0.71      & 0.81   & 0.76     & pcd      & 0.60      & 0.50   & 0.55     \\
    dty       & 0.51      & 0.65   & 0.57     & pdc      & 0.46      & 0.56   & 0.51     \\
    egl       & 0.94      & 0.84   & 0.89     & pfl      & 0.87      & 0.71   & 0.78     \\
    ell       & \textcolor{ForestGreen}{1.00} & 0.94   & 0.97     & pnb      & 0.76      & 0.97   & 0.86     \\
    eng       & 0.20      & 0.78   & 0.31     & pol      & 0.81      & 0.92   & 0.86     \\
    epo       & 0.56      & 0.71   & 0.62     & por      & 0.76      & 0.70   & 0.73     \\
    est       & 0.80      & 0.85   & 0.82     & pus      & 0.98      & 0.91   & 0.94     \\
    eus       & 0.98      & 0.95   & 0.96     & que      & 0.91      & 0.68   & 0.78     \\
    ext       & 0.64      & 0.66   & 0.65     & roa-tara & 0.80      & 0.91   & 0.85     \\
    fao       & 0.89      & 0.86   & 0.87     & roh      & 0.62      & 0.85   & 0.72     \\
    fas       & 0.50      & 0.84   & 0.63     & ron      & 0.89      & 0.83   & 0.86     \\
    fin       & 0.84      & 0.95   & 0.89     & rue      & 0.80      & 0.82   & 0.81     \\
    fra       & 0.50      & 0.72   & 0.59     & rup      & 0.92      & 0.78   & 0.84     \\
    frp       & 0.88      & 0.60   & 0.71     & rus      & 0.44      & 0.88   & 0.59     \\
    fry       & 0.89      & 0.80   & 0.84     & sah      & 0.97      & 0.90   & 0.94     \\
    fur       & 0.83      & 0.81   & 0.82     & san      & 0.94      & 0.96   & 0.95     \\
    gag       & 0.82      & 0.71   & 0.76     & scn      & 0.81      & 0.74   & 0.77     \\
    gla       & 0.95      & 0.94   & 0.94     & sco      & 0.42      & 0.28   & 0.34     \\
    gle       & 0.96      & 0.92   & 0.94     & sgs      & \textcolor{ForestGreen}{1.00} & 0.95   & 0.98     \\
    glg       & 0.57      & 0.68   & 0.62     & sin      & \textcolor{ForestGreen}{1.00} & 0.96   & 0.98     \\
    glk       & 0.87      & \textcolor{red}{0.19}   & 0.31     & slk      & 0.78      & 0.85   & 0.81     \\
    glv       & 0.92      & 0.96   & 0.94     & slv      & 0.59      & 0.74   & 0.66     \\
    grn       & 0.96      & 0.91   & 0.94     & sme      & 0.94      & 0.72   & 0.81     \\
    guj       & \textcolor{ForestGreen}{1.00} & 0.99   & 0.99     & sna      & 0.91      & 0.87   & 0.89     \\
    hak       & \textcolor{ForestGreen}{1.00} & 0.90   & 0.95     & snd      & 0.88      & 0.97   & 0.92     \\
    hat       & 0.95      & 0.93   & 0.94     & som      & 0.96      & 0.94   & 0.95     \\
    hau       & 0.88      & 0.92   & 0.90     & spa      & 0.42      & 0.50   & 0.46     \\
    hbs       & 0.39      & 0.08   & \textcolor{red}{0.14}     & sqi      & 0.99      & 0.90   & 0.94     \\
    heb       & 0.99      & \textcolor{ForestGreen}{1.00} & 0.99     & srd      & 0.75      & 0.84   & 0.79     \\
    hif       & 0.74      & 0.72   & 0.73     & srn      & 0.78      & 0.88   & 0.83     \\
    hin       & 0.64      & 0.82   & 0.72     & srp      & 0.30      & \textcolor{red}{0.12}   & \textcolor{red}{0.17}     \\
    hrv       & 0.32      & 0.56   & 0.41     & stq      & 0.94      & 0.78   & 0.85     \\
    hsb       & 0.74      & 0.56   & 0.64     & sun      & 0.69      & 0.32   & 0.44     \\
    hun       & 0.99      & 0.96   & 0.98     & swa      & 0.91      & 0.83   & 0.87     \\
    hye       & \textcolor{ForestGreen}{1.00} & 0.99   & 0.99     & swe      & 0.85      & 0.77   & 0.81     \\
    \end{tabular}
    \caption{Results of the character distribution (IDO metric) applied to \dbNameVersion.}
    \label{table:char-distribution-ido-results-1}
\end{table}

\begin{table}
    \begin{tabular}{llll|llll}
    \toprule
    \textbf{Lang}  & \textbf{Prec} & \textbf{Recall} & $\mathbf{F}_1$ & \textbf{Lang}  & \textbf{Prec} & \textbf{Recall} & $\mathbf{F_1}$ \\\midrule
    ibo       & 0.74      & 0.71   & 0.73     & szl      & 0.96      & 0.83   & 0.89     \\
    ido       & 0.40      & 0.63   & 0.49     & tam      & \textcolor{ForestGreen}{1.00} & 0.98   & 0.99     \\
    ile       & 0.57      & 0.43   & 0.49     & tat      & 0.98      & 0.81   & 0.89     \\
    ilo       & 0.88      & 0.87   & 0.87     & tcy      & 0.95      & 0.92   & 0.94     \\
    ina       & 0.39      & 0.82   & 0.53     & tel      & \textcolor{ForestGreen}{1.00} & 0.97   & 0.98     \\
    ind       & 0.35      & 0.28   & 0.31     & tet      & 0.74      & 0.78   & 0.76     \\
    isl       & 0.91      & 0.87   & 0.89     & tgk      & 0.96      & 0.88   & 0.92     \\
    ita       & 0.41      & 0.77   & 0.54     & tgl      & 0.52      & 0.71   & 0.60     \\
    jam       & 0.81      & 0.90   & 0.85     & tha      & 0.94      & 0.99   & 0.96     \\
    jav       & 0.74      & 0.63   & 0.68     & ton      & 0.99      & 0.79   & 0.88     \\
    jbo       & 0.93      & 0.99   & 0.96     & tsn      & 0.75      & 0.93   & 0.83     \\
    jpn       & \textcolor{ForestGreen}{1.00} & 0.97   & 0.98     & tuk      & 0.99      & 0.97   & 0.98     \\
    kaa       & 0.99      & 0.91   & 0.95     & tur      & 0.60      & 0.87   & 0.71     \\
    kab       & 0.93      & 0.94   & 0.94     & tyv      & 0.93      & 0.88   & 0.91     \\
    kan       & 0.93      & 0.95   & 0.94     & udm      & 0.83      & 0.85   & 0.84     \\
    kat       & 0.93      & 0.97   & 0.95     & uig      & \textcolor{ForestGreen}{1.00} & 0.89   & 0.94     \\
    kaz       & \textcolor{ForestGreen}{1.00} & 0.96   & 0.98     & ukr      & 0.91      & 0.91   & 0.91     \\
    kbd       & \textcolor{ForestGreen}{1.00} & 0.95   & 0.97     & urd      & 0.96      & 0.66   & 0.78     \\
    khm       & \textcolor{ForestGreen}{1.00} & 0.80   & 0.89     & uzb      & 0.92      & 0.94   & 0.93     \\
    kin       & 0.90      & 0.78   & 0.83     & vec      & 0.88      & 0.70   & 0.78     \\
    kir       & 0.85      & 0.75   & 0.80     & vep      & 0.90      & 0.80   & 0.85     \\
    koi       & 0.77      & 0.43   & 0.55     & vie      & 0.91      & 0.83   & 0.87     \\
    kok       & 0.88      & \textcolor{red}{0.10}   & \textcolor{red}{0.19}     & vls      & 0.62      & 0.54   & 0.58     \\
    kom       & 0.66      & 0.66   & 0.66     & vol      & 0.86      & 0.89   & 0.87     \\
    kor       & 0.93      & 0.97   & 0.95     & vro      & 0.97      & 0.87   & 0.92     \\
    krc       & 0.70      & 0.93   & 0.80     & war      & 0.82      & 0.98   & 0.90     \\
    ksh       & 0.93      & 0.85   & 0.89     & wln      & 0.87      & 0.84   & 0.86     \\
    kur       & 0.93      & 0.93   & 0.93     & wol      & 0.93      & 0.92   & 0.93     \\
    lad       & 0.76      & 0.86   & 0.80     & wuu      & 0.95      & 0.75   & 0.84     \\
    lao       & \textcolor{ForestGreen}{1.00} & 0.76   & 0.86     & xho      & 0.96      & 0.94   & 0.95     \\
    lat       & 0.68      & 0.82   & 0.74     & xmf      & 0.98      & 0.91   & 0.95     \\
    lav       & 0.88      & 0.91   & 0.90     & yid      & \textcolor{ForestGreen}{1.00} & 0.99   & \textcolor{ForestGreen}{1.00} \\
    lez       & 0.78      & 0.94   & 0.85     & yor      & 0.78      & 0.49   & 0.60     \\
    lij       & 0.91      & 0.66   & 0.77     & zea      & 0.72      & 0.35   & 0.47     \\
    lim       & 0.74      & 0.62   & 0.67     & zh-yue   & 0.81      & 0.79   & 0.80     \\
    lin       & 0.98      & 0.80   & 0.88     & zho      & 0.92      & 0.71   & 0.80     \\
    lit       & 0.81      & 0.89   & 0.84     & ~        & ~         & ~      & ~        \\\hline
    MEAN      & 0.81      & 0.78   & 0.78     & ~        & ~         & ~      & ~        \\\bottomrule
    \end{tabular}
    \caption{Results of the character distribution (IDO metric) applied to \dbNameVersion.}
    \label{table:char-distribution-ido-results-2}
\end{table}

\clearpage
\onecolumn
\begin{figure}
    \includegraphics[width=\linewidth]{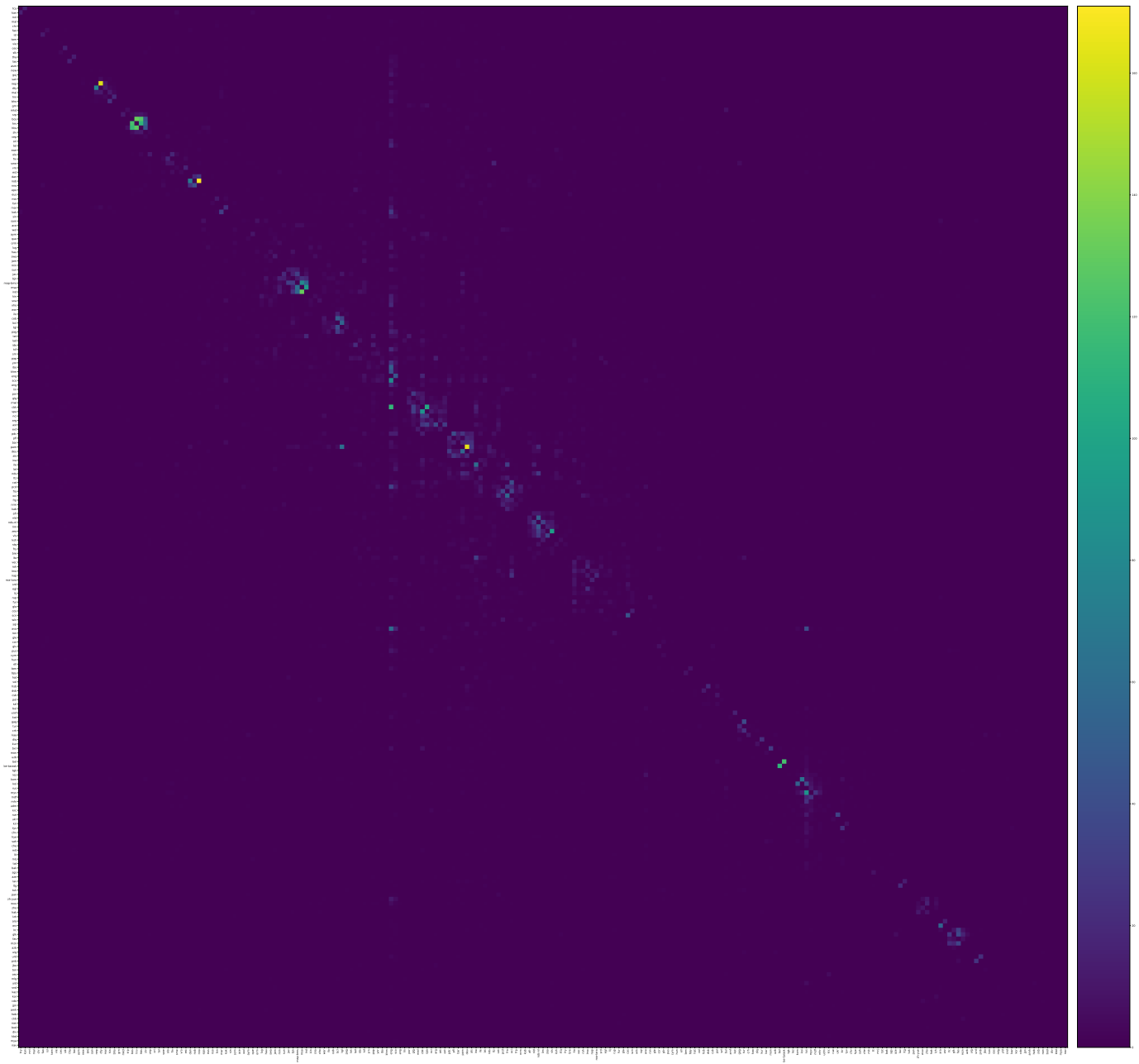}
    \caption{Confusion matrix of the tf-idf feature based neural network.
             The diagonal was set to 0 to emphazise the errors.
             The y-axis is the true language, the x-axis the predicted
             language. It shows that many languages get falsely predicted to be
             English.\\
             Best viewed digitally.}
    \label{fig:cm-tfidf-nn}
\end{figure}

\end{document}